# Can Perplexity Predict Finetuning Performance? An Investigation of Tokenization Effects on Sequential Language Models for Nepali

**Nishant Luitel, Nirajan Bekoju, Anand Kumar Sah and Subarna Shakya**
Dept. of Electronics and Computer Engineering,
Pulchowk Campus, Tribhuvan University,
Lalitpur, Nepal
{076bct041.nishant, 076bct039.nirajan, anand.sah}@pcampus.edu.np, drss@ioe.edu.np

## Abstract

The impact of subword tokenization on language model performance is well-documented for perplexity, with finer granularity consistently reducing this intrinsic metric. However, research on how different tokenization schemes affect a model's understanding capabilities remains limited, particularly for non-Latin script languages. Addressing this gap, we conducted a comprehensive evaluation of six distinct tokenization strategies by pretraining transformer-based language models for Nepali and evaluating their performance across multiple downstream tasks. While recent prominent models like GPT, RoBERTa, Claude, LLaMA, Mistral, Falcon, and MPT have adopted byte-level BPE tokenization, our findings demonstrate that for Nepali, SentencePiece tokenization consistently yields superior results on understanding-based tasks. Unlike previous studies that primarily focused on BERT-based architectures, our research specifically examines sequential transformer models, providing valuable insights for language model development in low-resource languages and highlighting the importance of tokenization strategy beyond perplexity reduction.

## 1 Introduction

Nepali, an Indo-Aryan language written in Devanagari script, serves as the official language of Nepal. According to the Nepal Population and Housing Census 2021, approximately 13 million people (44.9%) speak Nepali as their mother tongue, while an additional 13.5 million (46.2%) use it as their second language. The language extends beyond Nepal's borders into neighboring regions of India, Bhutan, Brunei, and Myanmar. Nepali follows a subject-object-verb sentence structure, distinguishing it from many Indo-European languages. Despite its significant speaker population, computational research in Nepali natural language processing remains underdeveloped due to limited high-quality datasets and computational resources. Nepali's rich morphological complexity and extensive vocabulary pose unique challenges for creating accurate and concise content. Investigating the applicability of state-of-the-art NLP technologies to Nepali not only benefits researchers and speakers but also has potential implications for other Devanagari-script languages such as Hindi, Sanskrit, Maithili, and Bhojpuri.

Tokenization—the process of segmenting text into smaller units such as words or subwords—forms the foundation of natural language processing pipelines. This critical preprocessing step enables computational systems to analyze and process human language by converting raw text into discrete units that algorithms can efficiently manipulate. The choice of tokenization strategy significantly impacts a model's ability to handle vocabulary coverage, out-of-vocabulary words, and morphological complexity. Recent advances in subword tokenization have revolutionized NLP by balancing vocabulary size constraints with linguistic flexibility, particularly for morphologically rich languages like Nepali.

Contemporary language models generate human-like text by leveraging transformer architectures trained on massive text corpora. These models primarily follow two paradigms: masked language modeling (MLM), exemplified by BERT (Devlin et al., 2018), where models learn bidirectional context by predicting masked tokens; and autoregressive language modeling, implemented in models like GPT (Radford et al., 2019; Brown et al., 2020) and PaLM (Chowdhery et al., 2022), where models predict the next token based on preceding context. While masked language models excel at learning powerful bidirectional representations suitable for downstream tasks, autoregressive models offer superior capabilities for text generation. Unlike previous studies that predominantly focused on BERT-based architec-tures for Nepali, our work specifically examines sequential (autoregressive) transformer models similar to (Luitel et al., 2024), trained with various tokenization strategies and evaluated on multiple downstream tasks.

The major contributions of our paper are as follows:

1. We pretrained 7 sequential language models using diverse tokenizers: Word Tokenizer (30,000 and 60,507 vocabs), SentencePiece, WordPiece, BPE, Morpheme, and Morpheme+BPE combination (all with 30,000 vocabs except as noted).

2. We compared language model performance based on perplexity during pre-training across different tokenization methods.

3. We evaluated pre-trained models by finetuning on multiple Nepali Natural Language Understanding (NLU) tasks and all code and models'll be made public on acceptance.

## 2 Related Works

Language modeling fundamentally aims to predict the next word given contextual words. Bengio et al. (2000) introduced the Neural Probabilistic Language Model (NPLM), which learns distributed word representations alongside probability functions for word sequences. Before Recurrent Neural Networks (RNNs) gained prominence, approaches based on parse trees and n-gram statistics dominated the field. Mikolov et al. (2010) demonstrated the superiority of RNN-based language models over standard n-gram techniques in speech recognition applications, despite their substantial computational complexity. Building on this foundation, Sutskever et al. (2011) advanced character-level modeling for text generation by training RNNs with the Hessian-Free optimizer. The field was revolutionized by Vaswani et al. (2017) with the introduction of the Transformer architecture, which implemented attention mechanisms to develop state-of-the-art machine translation models capable of generating text in one language given context in another. The Transformer's parallelization capabilities effectively addressed the computational and training limitations of previous sequential models, leading to the development of influential architectures like BERT (Devlin et al., 2018) and GPT (Brown et al., 2020) that now underpin numerous contemporary NLP tasks.

Recent years have witnessed growing research interest in pretraining and finetuning NLP models for low-resource languages like Nepali. Maskey (2023) pretrained a text generation model following Sanh et al. (2019)'s configuration on a combined dataset comprising Oscar, cc100, and scraped Nepali Wikipedia articles, employing SentencePiece tokenization with a 24,576 vocabulary size. Maskey et al. (2022) trained three distinct transformer-based masked language models (DistilBERT-base, DeBERTa-base, and XLM-RoBERTa) for Nepali text sequences, evaluating and comparing them against other transformer-based models on downstream classification tasks. In parallel work, Niraula and Chapagain (2022) finetuned Multilingual BERT specifically for Named Entity Recognition tasks in Nepali. Timilsina et al. (2022) developed another BERT-based language model for Nepali using WordPiece vocabulary with 30,522 subword tokens, demonstrating superior performance compared to other BERT-based language models (Rajan, 2021; Devlin et al., 2018; Conneau et al., 2020) when finetuned on four distinct tasks: Content Classification, Named Entity Recognition, Part-of-Speech Tagging, and Categorical Pair Similarity. Despite these various pretraining and finetuning efforts in Nepali, a comparative analysis of language model performance on downstream tasks using different tokenization approaches remains unexplored.

Several studies have investigated tokenization impacts in other languages. Toraman et al. (2022) analyzed the efficiency (training time, carbon emissions) and effectiveness (performance) of various tokenization techniques by finetuning a Turkish BERT-based language model on multiple downstream NLP tasks, finding that for similar and smaller vocabulary sizes, character-level BPE and WordPiece outperformed other approaches like word-based tokenization. For Korean, Park et al. (2020) discovered that morpheme tokenization followed by character-level BPE achieved optimal performance, as this approach prevents BPE from considering byte sequences spanning multiple morphemes. Alrefaie et al. (2024) observed similar results for Arabic, where combining BPE with morpheme-based approaches proved most effective. Additionally, Alyafeai et al. (2021) evaluated different tokenization methods on three Arabic NLP classification tasks, though without employing transformer-based architectures.

Our approach differs from these previous stud-ies in three significant ways. First, we finetune sequential (autoregressive) language models rather than BERT-based architectures. Second, we specifically analyze the performance of byte-level BPE tokenization algorithms—an aspect not thoroughly examined in prior work. Finally, we provide empirical evidence challenging the predictive validity of perplexity—the commonly used intrinsic metric during language model pretraining—regarding downstream finetuning performance.

## 3 Methodology

### 3.1 Tokenization Techniques

We have trained 6 different tokenizers keeping the vocabulary size at the constant of 30000. We intend to perform a comparison of LMs(perplexity and finetuning performance) but the perplexity scores tend to decrease with decreasing vocabulary size. Hence comparison through constant vocab size across models makes more sense. The table 1 shows encoded text for the same input by every tokenizer. Below are the specifics of how we trained these tokenizers.

1. **Word-based**: In our word-based tokenization scheme, we selected the top 30,000 vocabulary tokens based on frequency distribution. To handle out-of-vocabulary (OOV) words during training and evaluation, we incorporated a <unk>token. Additionally, we included a <num>token to efficiently encode all numerical strings in Nepali. We utilized PyTorch's torchtext library to construct this vocabulary.

2. **Morphemes**: Morphemes represent the smallest meaningful subdivisions of words. We employed the Morfessor 2.0 library to train a model that segments compound words into constituent morphemes using Maximum A Posteriori (MAP) estimation (Smit et al., 2014). This morfessor model was applied to approximately one-third of the OSCAR corpus to prepare a morpheme-level training dataset. Following the approach suggested by Park et al. (2020), we introduced a '*' token to indicate space between words, facilitating accurate reconstruction during decoding. Under this scheme, the text 'AB C' would be segmented as 'A B * C', preserving both morphological structure and word boundaries.

3. **WordPiece**: The WordPiece algorithm divides words into frequently occurring subword units. It initializes by segmenting words into characters and prepending '##' to non-initial tokens. For example, 'जीवन' would initially be segmented as '(ज, ##ी, ##व, ##न)'. The algorithm then combines these units based on the scoring function in equation 1, where '$f$' represents frequency:

$$score = \frac{f_{pair}}{f_{1st} * f_{2nd}} \tag{1}$$

This scoring mechanism prioritizes frequent combinations of infrequent subtokens. During encoding, WordPiece identifies the longest subtoken present in the vocabulary. We implemented this tokenizer using the 'Tokenizers' Python package, addressing compatibility issues with Devanagari diacritics by temporarily replacing them with English letters during preprocessing and reversing this substitution during decoding.

4. **SentencePiece(with BPE)**: For this tokenizer, we implemented character-level Byte Pair Encoding (BPE) compatible with SentencePiece. Unlike WordPiece, the BPE algorithm merges characters or subtokens based directly on merged token frequency, applying learned rules sequentially during encoding (Sennrich et al., 2016). Our implementation incorporates the white space handling capabilities introduced by Kudo and Richardson (2018), treating spaces as standard tokens rather than special delimiters. This approach was implemented using the 'Tokenizers' Python package.

5. **Byte-Level BPE**: Byte-level BPE operates similarly to character-level BPE but performs merging operations on individual bytes rather than characters. This approach provides stronger guarantees against OOV words by operating at a lower level of abstraction. However, byte-level BPE typically produces larger token sequences than character-level approaches for equivalent text, potentially affecting computational efficiency. The byte-level approach is particularly valuable for handling multilingual text and special characters.

6. **Morphemes and BPE**: In our final approach, we applied Morphemes and byte-level BPE

| Tokenization Method | Tokens |
|---|---|
| Word | [ 'महानायक', 'राजेश', 'हमाल', 'अहिले', 'चलचित्र', 'क्षेत्रमा', 'पातलिए', '।' ] |
| Morpheme | [ 'महानायक', '*', 'राजेश', '*', 'हमाल', '*', 'अहिले', '*', 'चलचित्र', '*', 'क्षेत्रमा', '*', 'पातलिए', '*', '।' ] |
| WordPiece | [ 'महान', '##ा', '##यक', 'राज□श', 'हमाल', 'अहिल□', 'चलचित□र', 'क□ष□त□रमा', 'पात', '##लिए', '।' ] |
| SentencePiece | [ '_मह', 'ानायक', '_राजेश', '_हमाल', '_अहिले', '_चलचित्र', '_क्षेत्रमा', '_पात', 'लिए', '_।' ] |
| BPE | **['à¤®à¤¹', 'à¤¾', 'à¤¨', 'à¤¾', 'à¤¯à¤ķ', ... 37 gibberish tokens]** |
| Mprpheme+ BPE | **['à¤®à¤¹', 'à¤¾', 'à¤¨', 'à¤¾', 'à¤¯à¤ķ', ... 37 gibberish tokens ]** |

Table 1: Comparison of tokenization methods for encoding the Nepali sentence 'महानायक राजेश हमाल अहिले चलचित्र क्षेत्रमा पातलिए ।'. The □ symbols in WordPiece tokenization represents an English letter used in place of one of the modifier character(diacritic).

tokenization algorithms sequentially. This combined method ensures that the resulting tokens do not span across morpheme boundaries, preserving linguistic structure while benefiting from BPE's compression capabilities. We applied byte-level BPE to the morpheme-segmented corpus created using the Morfessor library as described earlier, creating a tokenization scheme that respects both morphological and statistical patterns in the text.

### 3.2 Model Architecture

For every tokenization technique, the same model architecture was used for pretraining the language model. A simple architecture consisting of 6 layers of transformer encoder blocks with 6 attention heads each was modeled. The size of input embedding layer used for tokens was 300 and the dimension used for feedforeward network was 1024. To regularize, we used a dropout rate of 20%. Finally, both the batch size and the sequence length used were 64. The parameters used are summarized in the table 2. The total number of parameters in the 30k vocab LMs was 24M.

| Parameter | Value |
|---|---|
| emsize | 300 |
| dim_feedforeward | 1024 |
| nlayers | 6 |
| nhead | 6 |
| dropout | 0.2 |
| batch size | 64 |
| seq. length | 64 |

Table 2: Transformer Model Parameters

For finetuning, we added a hidden layer and an output layer feedforward network on top of the representation learned on the final layer of the last transformer block. The dimension of the hidden layer used was again 1024 with ReLU activation function, and the output layer's dimension was equal to the number of classes for the particular task.

## 4 Experiment

### 4.1 Dataset for LM Pre-training

We used Oscar corpus for the Nepali language (Ortiz Suárez et al., 2019) with the removal of duplicated sentences. The total data that became available from this corpus was 1.2GB. From this dataset, four versions of LMs were trained i.e. word-based, SentencePiece, WordPiece and BPE-tokenized LMs on 300k paragraphs while morphemes and morphemes with BPE-tokenized LMs were trained on 100k paragraphs. Before training the sentences were preprocessed, tokenized, encoded(given id), and then batched. After batching i.e. grouping 64 training examples, we get 16791 unique batches of training data when word-based tokenization is used. Using any other preprocessing and tokenization scheme led to larger number of batches as shown in Figure 1. The morpheme-based models were only trained on a third of the dataset hence the percentage was calculated relative to the batches calculated using word-based tokenization on this dataset.

### 4.2 Pre-Training

We trained 6 transformer-based language models using tokenizers of 3.1 with the architecture as described in 3.2. Additionally, we also trained a word-based language model with 60k vocabulary but the same model architecture. This provided us with some insights into performance based on vocabulary size. The model evaluation during the pertaining is based on the perplexity score which can be calculated using the eq. 2 where we have replaced $P(x_i|context)$ with $P(x_i)$.

$$\text{Perplexity} = \exp\left(-\frac{1}{N}\sum_{i=1}^{N}\log P(x_i)\right) \quad (2)$$

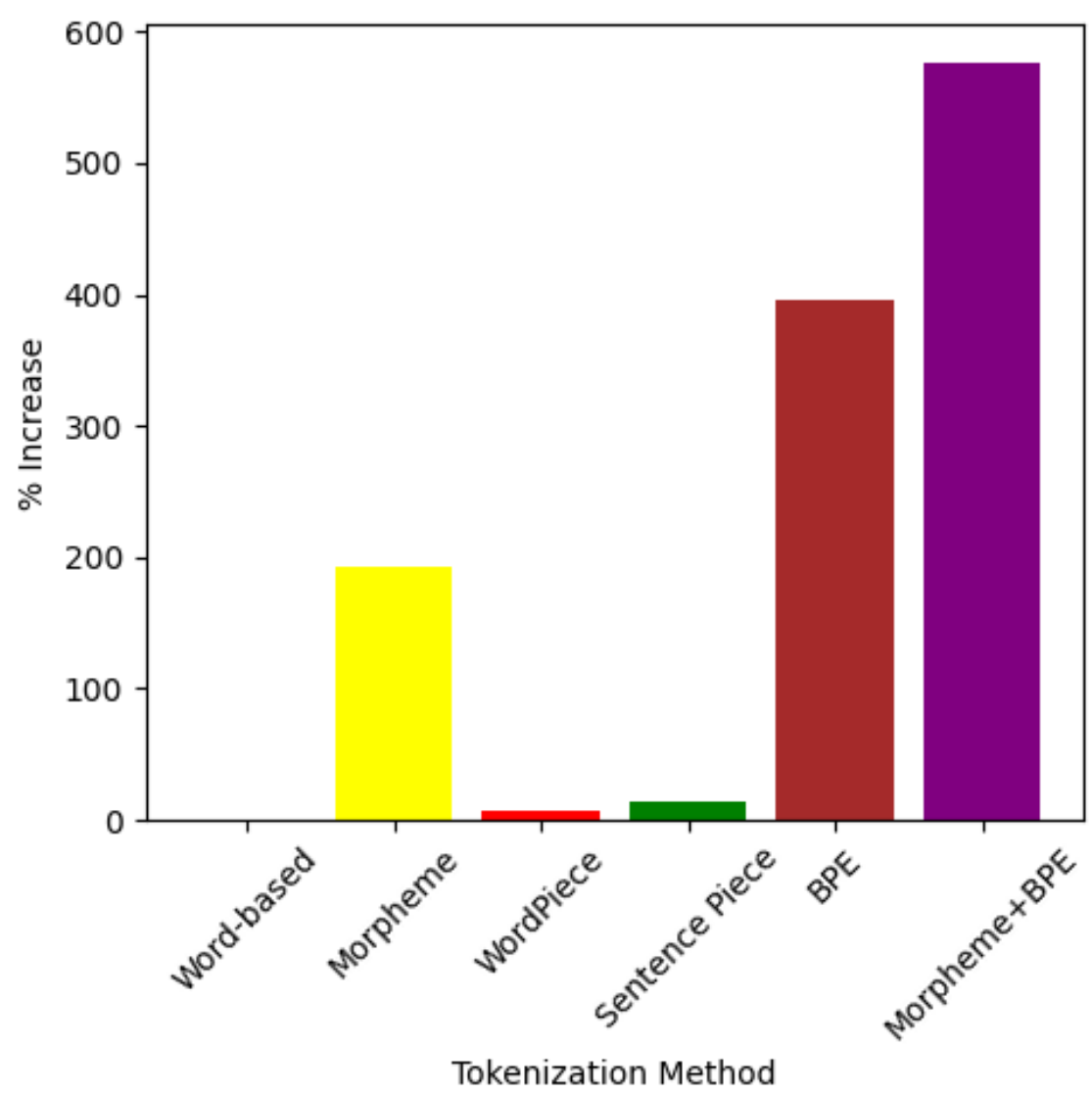

Figure 1: Percentage increase in number of batches with different tokenization methods relative to word-based tokenization.

### 4.3 Finetuning

The pre-trained language models were finetuned on Nep-gLUE benchmark datasets (Timilsina et al., 2022) which consists of four Natural Language Understanding tasks. The details on the finetuning approach and the datasets are briefly mentioned below:

#### 4.3.1 Categorical Pair Similarity(CPS)

Categorical Pair Similarity(CPS) is a pair-wise sequence classification task where the job is to find whether the given two sequences belong to the same category. CPS dataset was created (Timilsina et al., 2022) by extracting 2.5k of similar sequence pair for each of the 9 categories(total = 22.5k) and a 22.5k of different category sequence pair through random sampling accross dissimilar pair formed by pairing 2.5k sentences in each category with sentences from different category, resulting in a balanced dataset of 45k paired samples. Both of the sentences were passed through the pretrained model and the finetuning was performed on the concatenation of the representations from both the sequences. The prediction category was 1 for a similar pair and 0 for a dissimilar pair and, truncation was used whenever the sequence length limit was reached.

#### 4.3.2 Part of Speech Tagging(POS)

Part of Speech Tagging(POS) is a sequence labeling task where every word in the sequence of text has to be classified to one of tags such as noun, verb etc. This dataset was taken from a publicly available repository (Nepali Bhasa, 2020) which consists of 4251 sentences with more than 110k labels accross 39 tags. For preprocessing, multiple sequences for a same sentence was created and label was generted for each sequence. For example: Sentence ABC with words A(Tag: La),B(Tag: Lb) and C(Tag: Lc) can be decomposed into sentences A, AB, ABC. Then the label for sequence A is La , AB is Lb and ABC is Lc. Finally, the finetuning was performed using the representation of the last token. Hence to categorize the tag of B in sequence AB, we take the representation of B by passing AB into the pretrained model. Also, the truncation is performed from the beginning whenever the maximum sequence length is reached meaning that if the length limit is 2 then the sequence ABC would be trucated to BC.

#### 4.3.3 Named Entity Recognization(NER)

Similar to the POS task, Named Entity Recognition (NER) is also a sequence labeling task but here the job is to find the type of named entity like person, location or organization. The dataset used in the benchmark (Singh et al., 2019), consists around 3289 sentences with labels that belong to one of 7 classes including the other token '**O**'. Similar approach to POS tagging task was used as mentioned in sec. 4.3.2 in preprocessing, truncation and finetuning.

#### 4.3.4 Content Classification(CC)

Content classification is a task where the natural language content or sequence has to be classified in one of the categories. CC dataset was created (Timilsina et al., 2022) by scraping news articles from 9 different categories consisting of around 45k data points. The finetuning was performed on the sequence with truncation from the end.

## 5 Result and Discussion

### 5.1 Perplexity Trend

Table 3 shows the perplexity values at the end of training and validation. The training and validation perplexity is lowest for Morpheme with BPE followed by only BPE, while highest for SentencePiece followed by WordPiece. Notably, word-based tokenization outperforms both WordPiece and SentencePiece. Figure 2 illustrates the training and validation perplexity trends (in log scale) throughout training. All tokenization methods show initial steep decreases in training perplexity before flattening. Similarly, validation perplexity for WordPiece, SentencePiece, Word-level, and Morpheme shows large initial decreases before stabilizing. In contrast, byte-level BPE-based approaches display flat validation curves from the beginning, reflecting the large number of training steps already completed during the first epoch due to the higher number of batches processed when using byte-level tokenization.

| **Tokenization** | **Training** | **Validation** |
|---|---|---|
| BPE | 6.328 | 5.863 |
| Morpheme+BPE | 3.854 | 3.677 |
| SentencePiece | 134 | 120.6 |
| WordPiece | 125.6 | 116.3 |
| Morpheme | 14.09 | 13.71 |
| Word based(30k) | 106.8 | 97.08 |

Table 3: Perplexity values during training and validation

Figure 3 shows the comparison of the perplexity trend during training and validation for word-based tokenization with 30k tokens and 60k tokens. The perplexity score for 30k is less than for 60k during every phase of training and validation suggesting that an increase in vocab size in this region also tends to increase in perplexity.

### 5.2 Understanding Perplexity

| **Tokenization** | **% of most freq. token** |
|---|---|
| Morpheme+BPE | 0.160 |
| Bpe | 0.121 |
| SentencePiece | 0.047 |
| WordPiece | 0.168 |
| Morpheme | 0.479 |
| Word | 0.108 |

Table 4: Tokenization Methods and normalized frequency of the most frequent token

Our experiments reveal that tokenization methods involving Morpheme or BPE yield substantially lower perplexity scores compared to alternative approaches. This raises a critical question: Do these lower perplexity scores necessarily indicate superior language modeling capabilities? To investigate this relationship, we conducted a comprehensive frequency analysis on both training and evaluation corpora using the tokenizers trained on the training corpus, as illustrated in Figure 4.

The frequency distribution analysis across the entire vocabulary demonstrates that the SentencePiece algorithm maintains higher frequencies for mid-range tokens (up to the 25,000th token shown). We observe a clear correlation: tokenization methods yielding higher perplexity scores during evaluation consistently display higher frequency curves. However, examining the most frequent tokens—as shown in the frequency analysis of the top 15 vocabulary items—reveals that the SentencePiece algorithm, despite having the worst perplexity score, begins with the lowest normalized frequency. This pattern indicates that SentencePiece produces token distributions that are relatively more uniform compared to other algorithms evaluated in our study. This comparative uniformity suggests that when predicting the next token, models using SentencePiece assign less extreme probability to the most likely candidates. In practical terms, these models predict frequent tokens with less confidence while assigning relatively higher probabilities to less frequent tokens. Table 4 quantifies this difference dramatically: the most frequent token in SentencePiece covers only 4.7% of the corpus, while the most frequent token ('*') in the Morpheme approach spans 47.9% of the corpus. This explains why Morpheme tokenization achieves remarkably low perplexity—the model makes nearly half of its predictions with very high confidence.

From another perspective, BPE's superior perplexity performance stems from its ability to generate a larger number of high-frequency tokens compared to other methods. The byte-level BPE tokenization exhibits significantly higher normalized frequencies for approximately the first hundred most frequent tokens. Operating at the byte level rather than character level allows BPE to more efficiently capture repetitive patterns in text, leading to more confident predictions. However, this raises a fundamental question: Does this apparent advantage in perplexity metrics translate to enhanced understanding capacity?

Contrary to what perplexity scores might suggest, our experiments demonstrate that SentencePiece, the algorithm that performs worst according to perplexity standards, consistently outperforms other approaches when fine-tuned on natural language understanding (NLU) tasks. Additionally, despite their impressive perplexity scores, byte-level tokenization methods incur substantially higher computational costs during training. This inefficiency stems from their tendency to seg-

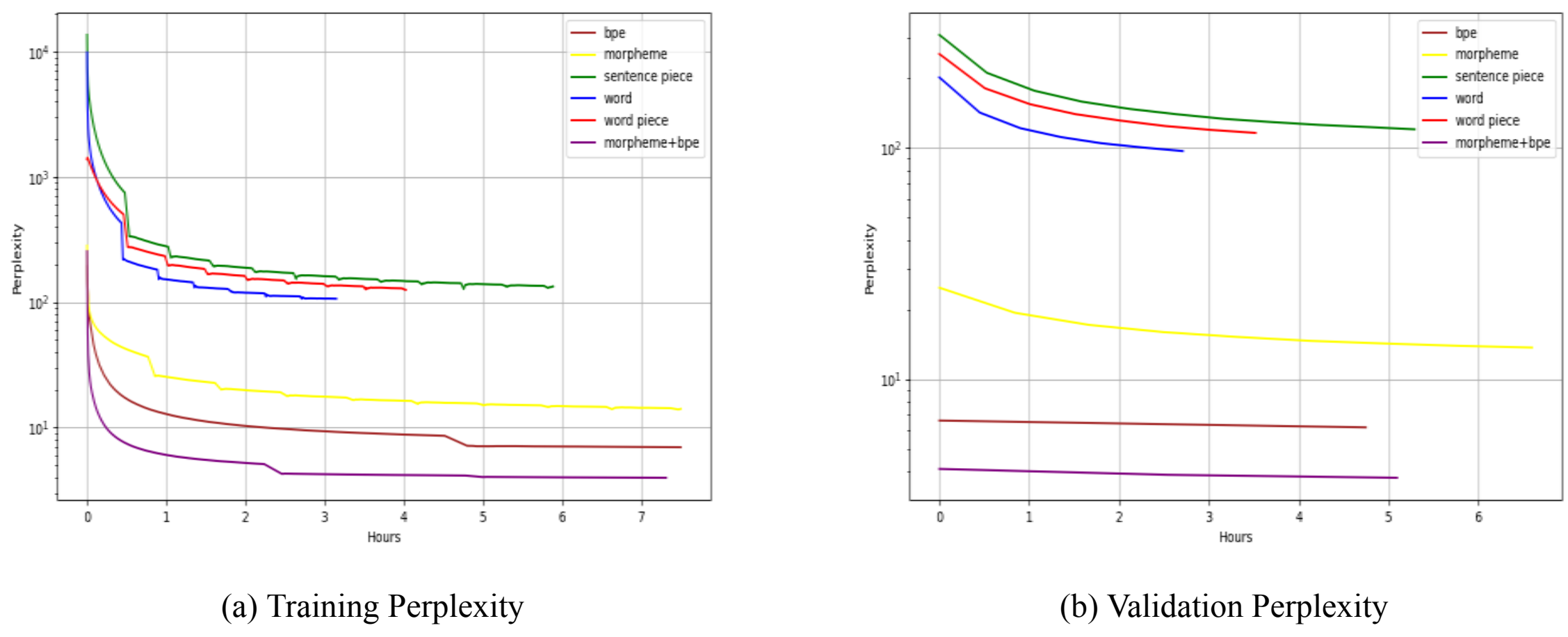

(a) Training Perplexity

(b) Validation Perplexity

Figure 2: Comparison of tokenization methods for perplexity

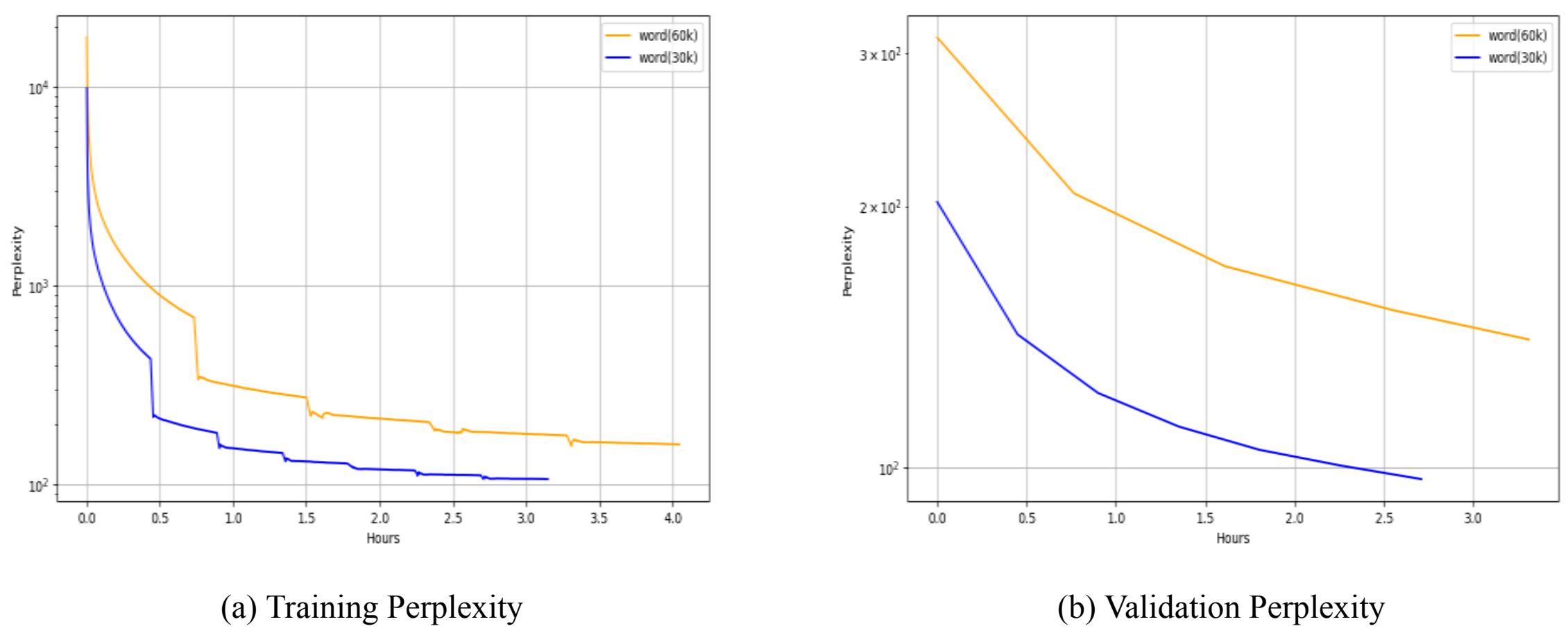

(a) Training Perplexity

(b) Validation Perplexity

Figure 3: Comparison of vocabulary size for perplexity

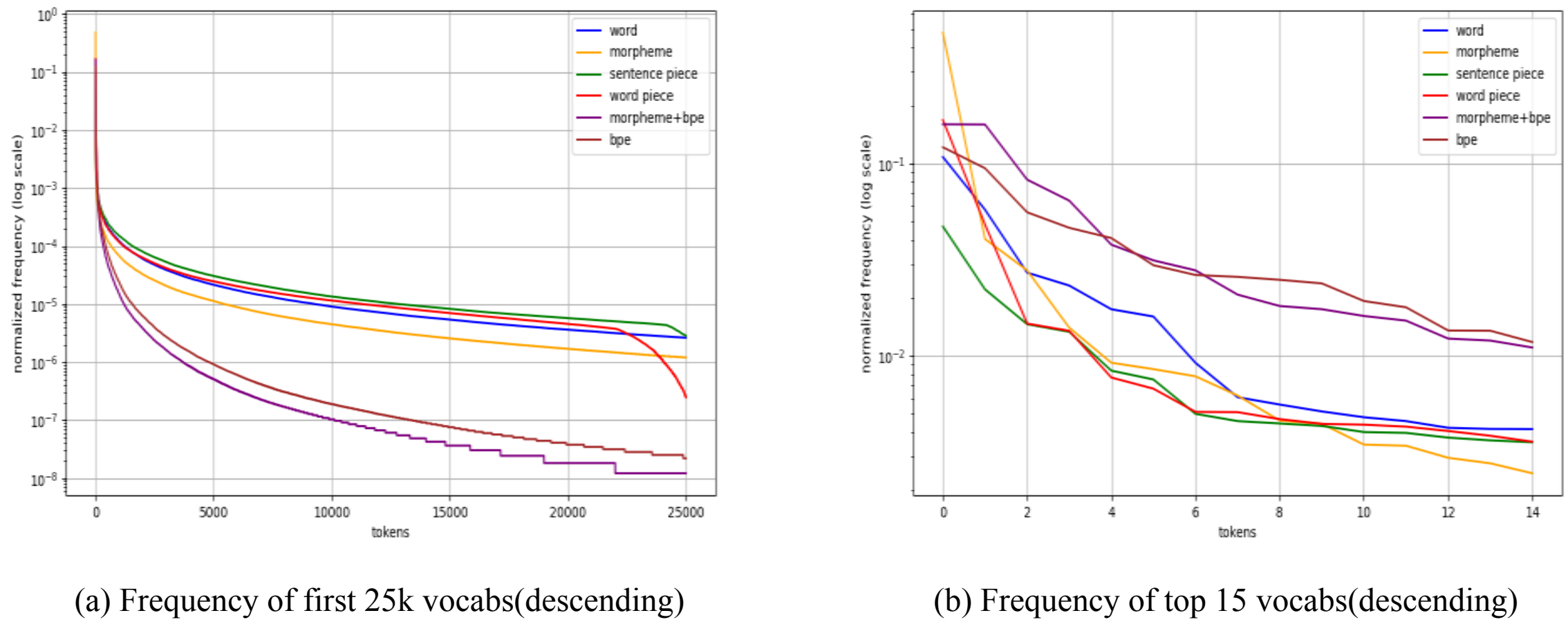

(a) Frequency of first 25k vocabs(descending)

(b) Frequency of top 15 vocabs(descending)

Figure 4: Comparison of normalized frequency of tokens in the corpus

| Tokenization | CPS | POS | NER | CC | NepGLUE |
|---|---|---|---|---|---|
| Morpheme+BPE | 0.86 | **0.90** | 0.72 | 0.77 | 0.81 |
| BPE | 0.89 | 0.87 | 0.75 | 0.81 | 0.83 |
| SentencePiece | **0.96** | 0.89 | 0.74 | **0.91** | **0.88** |
| WordPiece | 0.93 | 0.71 | 0.64 | 0.85 | 0.78 |
| Morpheme | 0.94 | 0.74 | **0.76** | 0.88 | 0.83 |
| Word (30k) | **0.96** | 0.75 | 0.72 | 0.90 | 0.83 |
| Word (60k) | **0.96** | 0.76 | 0.74 | **0.91** | 0.84 |

Table 5: Finetuning performance(Macro-F1 score) of language models with different tokenization schemes on four different NLU tasks Categorical Pair Similarity(CPS), Parts Of Speech Tagging(POS), Named Entity Recognition(NER) and Content Classification(CC) from Nep-gLUE benchmark. The final NepGLUE score represents the average performance across all tasks.

ment text into smaller token sequences, generating a larger total number of tokens during encoding. Beyond computational considerations, processing text as longer sequences of smaller tokens may impair contextual understanding when working with fixed sequence length limitations.

### 5.3 Finetuning Performance

Table 5 presents the results of finetuning on four tasks from the Nep-gLUE benchmark. The best-performing model for each task and the overall GLUE scores are highlighted in bold. Our analysis reveals several counterintuitive patterns regarding the relationship between perplexity and downstream performance.

For the Categorical Pair Similarity (CPS) task, SentencePiece—the worst-performing tokenization method in terms of perplexity—achieves the best macro-F1 score, tied with both 30k and 60k versions of word-based tokenization. Conversely, Morpheme+BPE, which demonstrated the lowest perplexity during pretraining, performs worst on this task. In Part-of-Speech (POS) tagging, Morpheme+BPE achieves the best macro-F1 score. However, SentencePiece, despite having the highest perplexity, outperforms all other tokenization methods except Morpheme+BPE. This finding further reinforces that perplexity is a poor predictor of a language model's representation learning capabilities.

For Named Entity Recognition (NER), the Morpheme algorithm performs best, with all other methods showing comparable performance except WordPiece, which performs significantly worse. In Content Classification (CC), SentencePiece again demonstrates superior performance, followed by word-based and Morpheme-based tokenization schemes, while byte-based algorithms perform considerably worse.

The averaged NepGLUE score across all tasks reveals that SentencePiece is the optimal tokenization method with a score of 0.88, while WordPiece performs worst with 0.78, followed by Morpheme+BPE with 0.81. This aligns with Liu et al. (2019)'s observations that byte-level BPE algorithms typically underperform compared to character-level BPE. Comparing word-based algorithms with 30k versus 60k vocabulary sizes, we observe that larger vocabulary size leads to marginally better or equivalent performance across tasks, without dramatic improvements. Unlike Toraman et al. (2022), we maintained consistent model sizes across different vocabulary sizes, which may explain the modest performance differences, as noted in Alrefaie et al. (2024).

## 6 Conclusion

In this paper, we compared perplexity scores across different tokenization methods using autoregressive language models for Nepali. We found that more granular tokenization typically produces fewer high-frequency tokens, resulting in lower perplexity. Increasing vocabulary size in word-based tokenization correspondingly increased perplexity. However, our finetuning experiments on various NLU tasks revealed that tokenization methods with the best perplexity scores (byte-level BPE with/without Morphemes) did not yield superior performance on understanding tasks. Instead, SentencePiece consistently outperformed other methods across tasks despite having worse perplexity scores.

## 7 Limitations

Despite our efforts, several limitations remain in this study. Our language models have only 24M parameters (30k versions), making them larger than the smallest BERT models (14M) but far from large-scale sequential models. Thus, the applicability of our findings to LLMs remains uncertain. Additionally, our models use a maximum sequence length of 64, which may bias comparisons between tokenization algorithms like byte-level BPE and word-based approaches in terms of contextual information, though the comparison remains fair computationally.

Furthermore, our benchmark datasets lack sequence generation tasks such as text summarization, machine translation, and question answering, limiting the generalizability of our results to generative models. While we evaluate six tokenization schemes, we do not consider alternatives like n-gram characters, Unigram LM (Kudo, 2018), or sampling-based SentencePiece (Kudo and Richardson, 2018), which could enhance robustness. A more comprehensive study incorporating these methods, as well as an analysis of vocabulary size effects beyond word-based tokenization, remains for future work. Finally, exploring larger models across multiple languages presents an interesting direction for further research.